\documentclass[graybox]{svmult}

\usepackage{mathptmx}       
\usepackage{helvet}         
\usepackage{courier}        
\usepackage{type1cm}        
\usepackage{makeidx}         
\usepackage{graphicx}        
\graphicspath{{Figures/}}
\usepackage{multicol}        
\usepackage[bottom]{footmisc}

\usepackage{cite}
\usepackage{multirow}
\usepackage{csvsimple}
\usepackage{array,booktabs}
\usepackage{amsmath}
\usepackage{dsfont}
\usepackage{amssymb}
\usepackage{cleveref}
\crefname{figure}{Fig.}{Figs.}
\crefname{table}{Table}{Tables}

\begin{document}

\title*{RSL19BD at DBDC4:\\Ensemble of Decision Tree-based and LSTM-based Models}
\titlerunning{RSL19BD at DBDC4}
\author{Chih-Hao Wang, Sosuke Kato, and Tetsuya Sakai}
\authorrunning{C.H. Wang et al.}
\institute{Chih-Hao Wang \at Waseda University, Shinjuku-ku Okubo 3-4-1 Tokyo Japan, \\ \email{haohaowang.oscar@moegi.waseda.jp}
\and Sosuke Kato \at Waseda University, Shinjuku-ku Okubo 3-4-1 Tokyo Japan, \email{sow@suou.waseda.jp}
\and Tetsuya Sakai \at Waseda University, Shinjuku-ku Okubo 3-4-1 Tokyo Japan, \email{tetsuyasakai@acm.org}}
%
%
\maketitle

\abstract*{RSL19BD (Waseda University Sakai Laboratory) participated in the Fourth Dialogue Breakdown Detection Challenge (DBDC4) and submitted five runs to both English and
Japanese subtasks. In these runs, we utilise the Decision Tree-based model and the Long Short-Term Memory-based (LSTM-based) model following the approaches of RSL17BD and KTH in the Third Dialogue Breakdown Detection Challenge (DBDC3) respectively.
The Decision Tree-based model follows the approach of RSL17BD but utilises RandomForestRegressor instead of ExtraTreesRegressor. In addition, instead of predicting the mean and the variance of the probability distribution of the three breakdown labels, it predicts the probability of each label directly. The LSTM-based model follows the approach of KTH with some changes in the architecture and utilises Convolutional Neural Network (CNN) to perform text feature extraction. In addition, instead of targeting the single breakdown label and minimising the categorical cross entropy loss, it targets the probability distribution of the three breakdown labels and minimises the mean squared error.
Run 1 utilises a Decision Tree-based model; 
Run 2 utilises an LSTM-based model;
Run 3 performs an ensemble of 5 LSTM-based models;
Run 4 performs an ensemble of Run 1 and Run 2;
Run 5 performs an ensemble of Run 1 and Run 3.
Run 5 statistically significantly outperformed all other runs in terms of MSE (NB, PB, B) for the English data and all other runs except Run 4 in terms of MSE (NB, PB, B) for the Japanese data
(alpha level = 0.05).}

\abstract{RSL19BD (Waseda University Sakai Laboratory) participated in the Fourth Dialogue Breakdown Detection Challenge (DBDC4) and submitted five runs to both English and
Japanese subtasks. In these runs, we utilise the Decision Tree-based model and the Long Short-Term Memory-based (LSTM-based) model following the approaches of RSL17BD and KTH in the Third Dialogue Breakdown Detection Challenge (DBDC3) respectively.
The Decision Tree-based model follows the approach of RSL17BD but utilises RandomForestRegressor instead of ExtraTreesRegressor. In addition, instead of predicting the mean and the variance of the probability distribution of the three breakdown labels, it predicts the probability of each label directly. The LSTM-based model follows the approach of KTH with some changes in the architecture and utilises Convolutional Neural Network (CNN) to perform text feature extraction. In addition, instead of targeting the single breakdown label and minimising the categorical cross entropy loss, it targets the probability distribution of the three breakdown labels and minimises the mean squared error.
Run 1 utilises a Decision Tree-based model; 
Run 2 utilises an LSTM-based model;
Run 3 performs an ensemble of 5 LSTM-based models;
Run 4 performs an ensemble of Run 1 and Run 2;
Run 5 performs an ensemble of Run 1 and Run 3.
Run 5 statistically significantly outperformed all other runs in terms of MSE (NB, PB, B) for the English data and all other runs except Run 4 in terms of MSE (NB, PB, B) for the Japanese data
(alpha level = 0.05).}

\section{Introduction}
\label{s:intro}

The task in the Fourth Dialogue Breakdown Detection Challenge (DBDC4)~\cite{DBDC4} is to build a model that detects whether an utterance from the system causes a breakdown in a dialogue context involving a system and a user. A breakdown is defined as a situation where the user cannot proceed with the conversation. Given a system utterance, the model is required to produce two outputs: 1. A single breakdown label chosen from the three breakdown labels (NB: Not a breakdown, PB: Possible breakdown, and B: Breakdown). 2. The probability distribution of the three breakdown labels, which we refer as P(NB), P(PB), and P(B) hereinafter. For evaluating the model, the organisers adopted classification-related metrics and distribution-related metrics and put an emphasis on the mean squared error (MSE). A complete description of the challenge can be found in the DBDC4 overview paper~\cite{DBDC4}. 

RSL19BD (Waseda University Sakai Laboratory) participated in DBDC4 and submitted five runs to both English and
Japanese subtasks. In these runs, we utilise the Decision Tree-based model and the Long Short-Term Memory-based (LSTM-based) model following the approaches of RSL17BD~\cite{RSL17BD} and KTH~\cite{KTH} in the Third Dialogue Breakdown Detection Challenge (DBDC3)~\cite{DBDC3} respectively.


\section{Prior Art}
\label{s:prior}

At DBDC3~\cite{DBDC3}, RSL17BD~\cite{RSL17BD} and KTH~\cite{KTH} both submitted models which achieved high performances. This section briefly describes their approaches.

\subsection{RSL17BD at DBDC3}
The top-performing model of RSL17BD utilises ExtraTreesRegressor~\cite{extratree}~\footnote{\url{https://scikit-learn.org/stable/modules/generated/sklearn.ensemble.ExtraTreesRegressor.html}} and employed six features shown in \cref{t:rsl17bd_feature_table} based on pattern analysis to predict the mean and variance of the probability distribution of the breakdown labels for each target system utterance. The predicted mean and variance are then converted into the predicted probability distribution of the three breakdown labels. The single breakdown label is determined by choosing the label with the highest probability.

\begin{table}[ht]
    \caption{The six features employed by RSL17BD at DBDC3}
    \label{t:rsl17bd_feature_table}
    \centering
    \begin{tabular}{ p{11.5cm} } 
        \toprule
        Feature \\
        \midrule
        turn-index of the target utterance \\
        \hline
        length of the target utterance (number of characters) \\
        \hline
        length of the target utterance (number of terms) \\
        \hline
        keyword flags of the target utterance \\
        \hline
        term frequency vector similarities among the target system utterance, the immediately preceding user utterance, and the system utterance that immediately precedes that user utterance \\
        \hline
        word embedding vector similarities among the target system utterance, the immediately preceding user utterance, and the system utterance that immediately precedes that user utterance \\
        \bottomrule
    \end{tabular}
\end{table}

\subsection{KTH at DBDC3}
The top-performing model of KTH utilises Long Short-Term Memory (LSTM)~\cite{LSTM}. For the preprocessing of English data, it produces a sequence of 300 dimensional word embedding vectors for each utterance in every dialogue and take the average sum of the sequence to produce the final embedding for a single utterance. The number of turns in each dialogue is fixed to 20 by removing the first system utterance which has no annotations or removing the last user turn. This produces an embedded dialogue of 20 turns, with each turn represented by a single 300 dimensional utterance embedding. The embedded dialogue is then processed by 4 LSTM layers and a Dense layer to produce 4 outputs for each turn. The 4 outputs are P(NB), P(PB), P(B), and P(U), where P(U) refers to the probability of user turn. The reason for adding P(U) is that user turns are included in the embedded dialogue as well and need to be predicted with a label different from NB, PB, and B. The model is trained for 100 epochs using Adadelta\cite{adadelta} as its optimiser. During training, it targets the single breakdown label and aims to minimise the categorical cross entropy loss for each target system utterance. For Japanese data, KTH did not submit any runs.

\section{Description of DBDC4 Dataset}
The development and evaluation dataset given in DBDC4 contain two languages: English and Japanese. 

The English data consists of dialogues from a dialogue system named IRIS and six other dialogue systems (anonymised as Bot001 to Bot006) which participated in the conversational intelligence challenge. In this paper, Bot001 to Bot006 are treated as a single system referred as BOT. Each dialogue is composed of 20 or 21 turns of alternating system and user utterances, with 10 system utterances being labeled. The labeled system utterances are evaluated by 15 human annotators, where each annotator labels it with a breakdown label chosen from NB, PB, and B. 

The Japanese data consists of two types of dialogues. The first type comes from three dialogue systems named DCM, DIT, and IRS. Each dialogue is composed of 21 turns of alternating system and user utterances, with 11 system utterances being labeled. The second type is located under a folder named \texttt{dbd\_livecompe\_eval} and comes from five systems (IRS, MMK, MRK, TRF, and ZNK) which participated in a live competition held in Japan. Each dialogue is composed of 31 turns of alternating system and user utterances, with 16 system utterances being labeled. In the development data, all labeled system utterances are evaluated by 30 human annotators. In the evaluation data, labeled system utterances of the first type and second type are evaluated by 15 and 30 human annotators respectively. A complete description of the dataset can be found in the DBDC4 overview paper~\cite{DBDC4}.  

For each dialogue system in the development dataset, we calculated the average probability distribution of the three breakdown labels across all its labeled utterances. We did not do so for the evaluation dataset since it is unlabeled. \cref{t:dbdc4_en_dev_data_table,t:dbdc4_jp_dev_data_table,t:dbdc4_jp_compe_dev_data_table} show our calculated results along with other statistical information.

\begin{table}[ht]
    \caption{Statistics of DBDC4 English data }
    \label{t:dbdc4_en_dev_data_table}
    \begin{tabular}{ c|c|c|c|c|c|c } 
        \toprule
        System name & No. of dialogues & No. of turns & No. of annotators & NB & PB & B \\
        \midrule
        BOT (dev) & 168 & 20 or 21 & 15 & 38.1\% & 28.7\% & 33.2\% \\
        \hline
        IRIS (dev) & 43 & 21 & 15 & 30.0\% & 30.3\% & 39.6\% \\
        \hline
        BOT (eval) & 173 & 20 or 21 & 15 & - & - & - \\
        \hline
        IRIS (eval) & 27 & 21 & 15 & - & - & - \\
        \bottomrule
    \end{tabular}
\end{table}

\begin{table}[ht]
    \caption{Statistics of DBDC4 Japanese data from DCM, DIT, and IRS}
    \label{t:dbdc4_jp_dev_data_table}
    \begin{tabular}{ c|c|c|c|c|c|c } 
        \toprule
        System name & No. of dialogues & No. of turns & No. of annotators & NB & PB & B \\
        \midrule
        DCM (dev) & 350 & 21 & 30 & 42.2\% & 29.9\% & 27.9\% \\
        \hline
        DIT (dev) & 150 & 21 & 30 & 26.0\% & 29.6\% & 44.4\% \\
        \hline
        IRS (dev) & 150 & 21 & 30 & 30.5\% & 25.8\% & 43.7\% \\
        \hline
        DCM (eval) & 50 & 21 & 15 & - & - & - \\
        \hline
        DIT (eval) & 50 & 21 & 15 & - &- & - \\
        \hline
        IRS (eval) & 50 & 21 & 15 & - & - & - \\
        \bottomrule
    \end{tabular}
\end{table}

\clearpage

\begin{table}[ht]
    \caption{Statistics of DBDC4 Japanese data from the five dialogue systems under \texttt{dbd\_livecompe\_eval}}
    \label{t:dbdc4_jp_compe_dev_data_table}
    \begin{tabular}{ c|c|c|c|c|c|c } 
        \toprule
        System name & No. of dialogues & No. of turns & No. of annotators & NB & PB & B \\
        \midrule
        IRS (dev) & 13 & 31 & 30 & 32.8\% & 25.4\% & 41.7\% \\
        \hline
        MMK (dev) & 15 & 31 & 30 & 57.6\% & 29.4\% & 13.0\% \\
        \hline
        MRK (dev) & 15 & 31 & 30 & 48.5\% & 35.5\% & 16.0\% \\
        \hline
        TRF (dev) & 14 & 31 & 30 & 69.4\% & 20.0\% & 10.6\% \\
        \hline
        ZNK (dev) & 16 & 31 & 30 & 47.2\% & 30.6\% & 22.2\% \\
        \hline
        IRS (eval) & 15 & 31 & 30 & - & - & - \\
        \hline
        MMK (eval) & 14 & 31 & 30 & - & - & - \\
        \hline
        MRK (eval) & 14 & 31 & 30 & - & - & - \\
        \hline
        TRF (eval) & 16 & 31 & 30 & - & - & - \\
        \hline
        ZNK (eval) & 14 & 31 & 30 & - & - & - \\
        \bottomrule
    \end{tabular}
\end{table}

\section{Model Descriptions}
\label{s:model}

\subsection{Decision Tree-based model}
For the preprocessing of both English and Japanese data, we follow the same approach as RSL17BD~\cite{RSL17BD} at DBDC3~\cite{DBDC3}. Our model employs the same set of features as RSL17BD's model, but utilises RandomForestRegressor~\cite{randomforest}~\footnote{\url{https://scikit-learn.org/stable/modules/generated/sklearn.ensemble.RandomForestRegressor.html}} instead of ExtraTreesRegressor~\cite{extratree}. In addition, instead of predicting the mean and the variance of the probability distribution over the three breakdown labels and then deriving the probability of each label, it predicts the probability of each label directly. The probability distribution is then calculated by normalising the probability of the three labels by their sum. 

The modifications above are decided by training and evaluating different model configurations using the English~\footnote{When training and evaluating a model using the English data from DBDC3, we used the revised data mentioned in the DBDC3 overview paper~\cite{DBDC3}.} and Japanese data from DBDC3~\footnote{Due to the late release of dataset for DBDC4, we first built our models using the dataset from DBDC3.}. The evaluation results are shown in \cref{t:rsl17bd_dbdc3_en_table,t:rsl17bd_dbdc3_jp_table}. \textbf{DT} means the model utilises decision trees, \textbf{EX10} means the model utilises ExtraTreesRegressor with 10 estimators, and \textbf{RF100} means the model utilises RandomForestRegressor with 100 estimators. \textbf{AV} means the model predicts the mean and the variance of the probability distribution, and \textbf{NBPBB} means the model predicts the probability of each label directly. There are four evaluation metrics. Accuracy denotes the number of correctly predicted breakdown labels divided by the total number of breakdown labels to be predicted (the larger the better); F1 (B) denotes the F1-measure where only the B labels are considered correct (the larger the better); JSD (NB, PB, B) denotes the Jensen-Shannon Divergence between the predicted and correct probability distribution (the smaller the better); MSE (NB, PB, B) denotes the mean squared error between the predicted and correct probability distribution (the smaller the better). The results show that \textbf{DT-RF100-NBPBB} outperformed the model submitted by RSL17BD in DBDC3 (\textbf{DT-EX10-AV}) in all evaluation metrics in both English and Japanese data. Thus, we chose to utilise the configuration of \textbf{DT-RF100-NBPBB} when submitting the model for Run 1.

\begin{table}[ht]
    \caption{Results of Decision Tree-based model with different configurations on DBDC3 English evaluation data}
    \label{t:rsl17bd_dbdc3_en_table}
    \begin{tabular}{ c|c|c|c|c } 
        \toprule
        Model & Accuracy & F1 (B) & JSD (NB, PB, B) & MSE (NB, PB, B)\\
        \midrule
        DT-EX10-AV & 0.3430 & 0.2344 & 0.0594 & 0.0357\\
        \hline
        DT-EX10-NBPBB & 0.4065 & \textbf{0.3696} & 0.0498 & 0.0291\\
        \hline
        DT-RF10-NBPBB & 0.3950 & 0.3542 & 0.0486 & 0.0282 \\
        \hline
        \textbf{DT-RF100-NBPBB} & \textbf{0.4095} & 0.3548 & \textbf{0.0466} & \textbf{0.0271} \\
        \bottomrule
    \end{tabular}
\end{table}

\begin{table}[ht]
    \caption{Results of Decision Tree-based model with different configurations on DBDC3 Japanese evaluation data}
    \label{t:rsl17bd_dbdc3_jp_table}
    \begin{tabular}{ c|c|c|c|c } 
        \toprule
        Model & Accuracy & F1 (B) & JSD (NB, PB, B) & MSE (NB, PB, B)\\
        \midrule
        DT-EX10-AV & 0.3927 & 0.3225 & 0.1297 & 0.0769 \\
        \hline
        DT-EX10-NBPBB & 0.5303 & 0.6050 & 0.0920 & 0.0502 \\
        \hline
        DT-RF10-NBPBB & 0.5455 & 0.6292 & 0.0875 & 0.0481 \\
        \hline
        \textbf{DT-RF100-NBPBB} & \textbf{0.5630} & \textbf{0.6511} & \textbf{0.0845} & \textbf{0.0460} \\
        \bottomrule
    \end{tabular}
\end{table}

\subsection{LSTM-based model}
\begin{figure}[h]
    \centering
    \includegraphics[width=11cm]{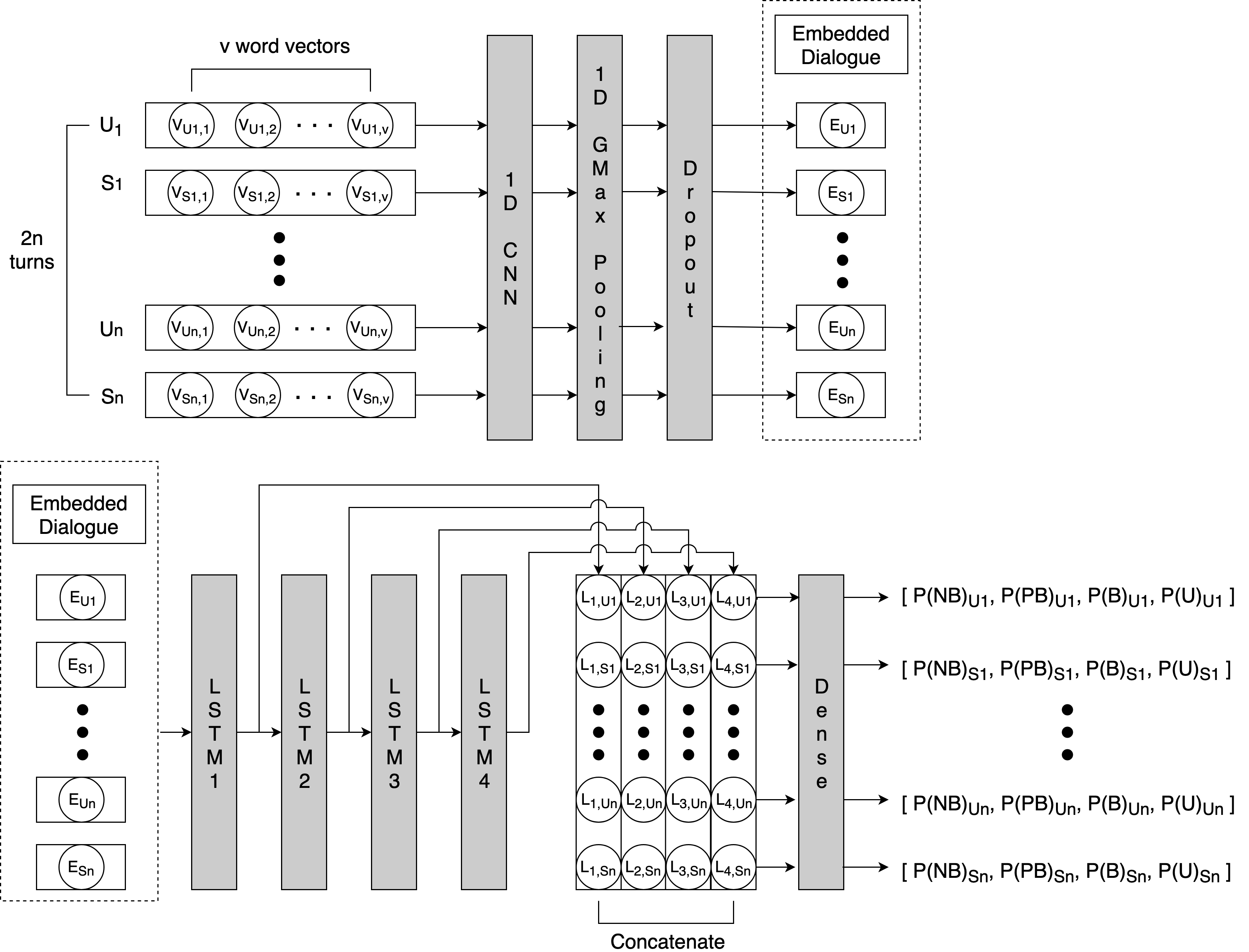}
    \caption{Architecture diagram of our LSTM-based model}
    \label{f:lstm_based_model}
\end{figure}

Following the approach of KTH~\cite{KTH} at DBDC3~\cite{DBDC3}, we utilise Long Short-Term Memory (LSTM)~\cite{LSTM}. However, instead of taking the average sum of word embedding vectors for each utterance, we utilise Convolutional Neural Networks (CNN) to perform text feature extraction and produce the final embedded utterance. In addition, instead of targeting the single breakdown label and minimising the categorical cross entropy loss for each target system utterance, our model targets the probability distribution of the three breakdown labels and minimises its mean squared error. We chose Adam~\cite{adam} as our optimiser and mean squared error as our loss function. \cref{f:lstm_based_model} shows the architecture diagram of our model.

For the preprocessing of both English and Japanese data, we first follow the same approach as RSL17BD at DBDC3 to produce a sequence of 300 dimensional word embedding vectors for each utterance in every dialogue. The number of word vectors in each sequence is fixed to $v$, with $v$ set to 50. This is done by truncating sequences that are longer than $v$ and padding sequences that are shorter than $v$ with zero vectors. The number of turns in a each dialogue is fixed to $2n$ by either removing the first system utterance which has no annotations or removing the last user turn. For the English data and the Japanese data from DCM, DIT, and IRS, the number of turns in each dialogue is fixed to 20 by setting $n$ to 10. For the data of the five dialogue systems under \texttt{dbd\_livecompe\_eval}, the number of turns in each dialogue is fixed to 30 by setting $n$ to 15.

The process above produces a dialogue of $2n$ turns, with each turn represented by a sequence of $v$ word vectors. We apply One-dimensional Convolutional Neural Networks (1D CNN), One-dimensional Global Max Pooling (1D GMax Pooling), and Dropout~\cite{Dropout} for each sequence to produce an embedded dialogue. The 1D CNN layer uses 150 filters of size 2 with ReLU~\cite{relu} as the activation function. The dropout rate of the Dropout layer is set to 0.4.

The embedded dialogue is then processed by 4 LSTM layers sequentially. Each LSTM layer contains 64 units, with dropout set to 0.1, and recurrent dropout set to 0.1. We used LSTM instead of Bi-LSTM because the usage of turns after the target system utterance is disallowed. The output sequences from the 4 LSTM layers are concatenated to form a ($2n$, 256) dimensional matrix, and processed by a Dense layer with softmax activation and 4 outputs. The 4 outputs represent P(NB), P(PB), P(B), and P(U) respectively. The probability distribution for each target system utterance is calculated by normalising P(NB), P(PB), and P(B) by their sum. The single breakdown label is determined by choosing the label with the highest probability in the distribution.

The modifications above are decided by training and evaluating different model configurations using the English and Japanese data from DBDC3. The evaluation results are shown in \cref{t:lstm_dbdc3_en_table,t:lstm_dbdc3_jp_table}. \textbf{LSTM} means the model utilises LSTM, and \textbf{LTSM-CNN} means the model utilises LSTM and CNN. \textbf{ADAD-CAT} means the model utilises Adadelta as optimizer and categorical cross entropy as loss function, and \textbf{ADAM-MSE} means the model utilises Adam as optimizer and mean squared error as loss function. 

The results show that for English data, \textbf{LSTM-CNN-ADAM-MSE} outperformed \textbf{LSTM-ADAD-CAT} and \textbf{LSTM-ADAM-MSE} in all evaluation metrics except F1 (B). Although \textbf{LSTM-ADAD-CAT} achieved high performance in F1 (B), its performance in mean squared error (MSE (NB, PB, B)) was poor. Since mean squared error is emphasised in this challenge, we decided to discard \textbf{LSTM-ADAD-CAT}. For Japanese data, \textbf{LSTM-CNN-ADAM-MSE} outperformed \textbf{LSTM-ADAM-MSE} in all evaluation metrics. We did not evaluate \textbf{LSTM-ADAD-CAT} because it is already discarded after the evaluation of English data. In the end, we chose to utilise the configuration of \textbf{LSTM-CNN-ADAM-MSE} when submitting the models for Run 2 and Run 3.

\begin{table}[ht]
    \caption{Results of LSTM-based model with different configurations on DBDC3 English evaluation data}
    \label{t:lstm_dbdc3_en_table}
    \begin{tabular}{ c|c|c|c|c|c } 
        \toprule
        Model & Epochs &  Accuracy & F1 (B) & JSD (NB, PB, B) & MSE (NB, PB, B) \\
        \midrule
        LSTM-ADAD-CAT & 100 & 0.4130 & \textbf{0.4616} & 0.0928 & 0.0573 \\
        \hline
        LSTM-ADAM-MSE & 100 & 0.3940 & 0.3714 & 0.0516 & 0.0300 \\
        \hline
        \textbf{LSTM-CNN-ADAM-MSE} & 50 & \textbf{0.4620} & 0.4268 & \textbf{0.0474} & \textbf{0.0274} \\
        \bottomrule
    \end{tabular}
\end{table}

\begin{table}[ht]
    \caption{Results of LSTM-based model with different configurations on DBDC3 Japanese evaluation data}
    \label{t:lstm_dbdc3_jp_table}
    \begin{tabular}{ c|c|c|c|c|c } 
        \toprule
        Model & Epochs &  Accuracy & F1 (B) & JSD (NB, PB, B) & MSE (NB, PB, B) \\
        \midrule
        LSTM-ADAM-MSE & 100 & 0.5448 & 0.6148 & 0.0885 & 0.0497 \\
        \hline
        \textbf{LSTM-CNN-ADAM-MSE} & 50 & \textbf{0.5739} & \textbf{0.6594} & \textbf{0.0826} & \textbf{0.0463} \\
        \bottomrule
    \end{tabular}
\end{table}


\section{Runs}
\label{s:runs}
The descriptions of our runs are shown in \cref{t:run_table}. 
In Runs 1-3, we used the same strategy in creating the training data from the given development data in DBDC4. 
For the English submission, we created a single group of training data which consists of the entire English development data. We refer it as $E_{t1}$ hereinafter. The entire English evaluation data is referred as $E_{e1}$ hereinafter.
For the Japanese submission, we created two groups of training data. The first group consists of the development data from DCM, DIT, and IRS, and the second group consists of the development data from the five dialogue systems under \texttt{dbd\_livecompe\_eval}. We refer them as $J_{t1}$ and $J_{t2}$ hereinafter. The evaluation data from DCM, DIT, and IRS and the evaluation data from the five dialogue systems under \texttt{dbd\_livecompe\_eval} are referred as $J_{e1}$ and $J_{e2}$ hereinafter.

\begin{table}[ht]
    \caption{Description of runs for English and Japanese}
    \label{t:run_table}
    \begin{tabular}{ c|p{10cm}} 
        \toprule
        Run & \multicolumn{1}{c}{Description}\\
        \midrule
        1 & Decision Tree-based model\\
        2 & LSTM-based model\\
        3 & Ensemble of 5 LSTM-based models \\
        4 & Ensemble of Run 1 and Run 2 \\
        5 & Ensemble of Run 1 and Run 3 \\
        \bottomrule
    \end{tabular}
\end{table}



\subsection{Run 1: Decision Tree-based model}

For the English submission, we trained our Decision Tree-based model with $E_{t1}$ and made predictions on $E_{e1}$. For the Japanese submission, we built two models by training one with $J_{t1}$, and the other with $J_{t2}$. We made predictions on $J_{e1}$ using the former model and $J_{e2}$ using the latter model.

\subsection{Run 2: LSTM-based model}

For the English submission, we pretrained our LSTM-based model for 30 epochs with the entire English development and evaluation data in DBDC3, fine-tuned it by training for 32 epochs with $E_{t1}$, and made predictions on $E_{e1}$. For the Japanese submission, we built two LSTM-based models. The first model is trained for 30 epochs with $J_{t1}$. The second model is created by loading the weights from the first model and fine-tuning for 25 epochs with $J_{t2}$. We made predictions on $J_{e1}$ using the first model and $J_{e2}$ using the second model. Every model is trained using a batch size of 32.

\subsection{Run 3: Ensemble of 5 LSTM-based models }
The way an ensemble of 5 LSTM-based models is built is described as follows: Given training data $D_{t}$ and evaluation data $D_{e}$, we randomly divide $D_{t}$ into 10 portions and sample 5 portions from it. We build 5 models, where each model is trained using one of the sampled portions as validation data and the rest of the development data as training data. The batch size is set to 32. Each model is saved when the validation loss is minimum and no overfitting occurred. We make predictions on $D_{e}$ using each model, and take the mean of the predicted probability distribution for each target system utterance from the 5 models to produce a new probability distribution. The new single breakdown label is determined by choosing the label with the highest probability in the new probability distribution.


For the English submission, we pretrained an LSTM-based model for 30 epochs with the entire English development and evaluation data in DBDC3. The ensemble of 5 LSTM-based models is built by fine-tuning the pretrained model with $D_{t}=E_{t1}$ and $D_{e}=E_{e1}$.
The results of each LSTM-based model on the sampled validation data are shown in \cref{t:lstm_dbdc4_en_table}. 



For the Japanese submission, we built two ensemble models. The first model is built with $D_{t}=J_{t1}$ and $D_{e}=J_{e1}$. The results of each LSTM-based model on the sampled validation data are shown in \cref{t:lstm_dbdc4_jp_table}. The second model is built by loading the weights of the first model from the Japanese submission in Run 2 and fine-tuning it with $D_{t}=J_{t2}$ and $D_{e}=J_{e2}$. The results of each LSTM-based model on the sampled validation data are shown in \cref{t:lstm_dbdc4_jp_compe_table}.

\begin{table}[ht]
    \caption{Results of each LSTM-based model in Run 3 on the sampled validation data from $E_{t1}$}
    \label{t:lstm_dbdc4_en_table}
    \begin{tabular}{ c|c|c|c|c } 
        \toprule
        Model & Accuracy & F1 (B) & JSD (NB, PB, B) & MSE (NB, PB, B) \\
        \midrule
        1 & 0.5286 & 0.5385 & 0.0649 & 0.0343 \\
        \hline
        2 & 0.5190 & 0.5318 & 0.0701 & 0.0370 \\
        \hline
        3 & 0.5524 & 0.5660 & 0.0713 & 0.0375 \\
        \hline
        4 & 0.5381 & 0.6306 & 0.0706 & 0.0370 \\
        \hline
        5 & 0.5810 & 0.6635 & 0.0771 & 0.0401 \\
        \bottomrule
    \end{tabular}
\end{table}

\begin{table}[ht]
    \caption{Results of each LSTM-based model in Run 3 on the sampled validation data from $J_{t1}$}
    \label{t:lstm_dbdc4_jp_table}
    \begin{tabular}{ c|c|c|c|c } 
        \toprule
        Model & Accuracy & F1 (B) & JSD (NB, PB, B) & MSE (NB, PB, B) \\
        \midrule
        1 & 0.5664 & 0.5782 & 0.0887 & 0.0469 \\
        \hline
        2 & 0.5804 & 0.6057 & 0.0914 & 0.0477 \\
        \hline
        3 & 0.5944 & 0.6505 & 0.0786 & 0.0429 \\
        \hline
        4 & 0.5944 & 0.6402 & 0.0903 & 0.0473 \\
        \hline
        5 & 0.5846 & 0.6231 & 0.0961 & 0.0495 \\
        \bottomrule
    \end{tabular}
\end{table}

\begin{table}[ht]
    \caption{Results of each LSTM-based model in Run 3 on the sampled validation data from $J_{t2}$}
    \label{t:lstm_dbdc4_jp_compe_table}
    \begin{tabular}{ c|c|c|c|c } 
        \toprule
        Model & Accuracy & F1 (B) & JSD (NB, PB, B) & MSE (NB, PB, B) \\
        \midrule
        1 & 0.6313 & 0.4583 & 0.0706 & 0.0371 \\
        \hline
        2 & 0.6953 & 0.4324 & 0.0591 & 0.0323 \\
        \hline
        3 & 0.6484 & 0.3750 & 0.0510 & 0.0277 \\
        \hline
        4 & 0.6797 & 0.3529 & 0.0600 & 0.0319 \\
        \hline
        5 & 0.6016 & 0.0000 & 0.0678 & 0.0336 \\
        \bottomrule
    \end{tabular}
\end{table}

\subsection{Run 4: Ensemble of Run 1 and Run 2}
For both English and Japanese submissions, we take the mean of the predicted probability distribution for each target system utterance from Run 1 and Run 2 to produce a new probability distribution. The new single breakdown label is determined by choosing the label with the highest probability in the new probability distribution.

\subsection{Run 5: Ensemble of Run 1 and Run 3}
This run is identical with Run 4 except that Run 2 is replaced by Run 3.

\section{Results}
\label{s:results}
\cref{t:dbdc4_result_en_table,t:dbdc4_result_jp_table} show the official results of our English and Japanese runs respectively. It can be observed that Run 5 did well on average. For English runs, it outperformed all other runs in all evaluation metrics. For Japanese runs, it outperformed all other runs in JSD (NB, PB, B) and MSE (NB, PB, B). 

\cref{t:dbdc4_tukey_mse_en_table,t:dbdc4_tukey_mse_jp_table} show the results of comparing the MSE (NB, PB, B) of Runs 1-5 based on the Randomised Tukey's Honestly Significant Differences (HSD) test. The test is conducted with 10,000 replicates. The p-values are shown alongside with effect sizes (standardised mean differences)~\cite{effectsize}. \cref{t:dbdc4_tukey_mse_en_table} shows that Run 5 statistically significantly outperformed all other runs in terms of MSE (NB, PB, B) for the English data. \cref{t:dbdc4_tukey_mse_jp_table} shows that Run 5 statistically significantly outperformed all other runs except Run 4 in terms of MSE (NB, PB, B) for the Japanese data. The p-values show that the differences are statistically significant at the alpha level of 0.05.

\begin{table}[ht]
    \caption{Official results on English data}
    \label{t:dbdc4_result_en_table}
    \begin{tabular}{ c|c|c|c|c } 
        \toprule
        Model & Accuracy & F1 (B) & JSD (NB, PB, B) & MSE (NB, PB, B) \\
        \midrule
        Run 1 & 0.4990 & 0.4411 & 0.0700 & 0.0362 \\
        \hline
        Run 2 & 0.4730 & 0.4483 & 0.0725 & 0.0374 \\
        \hline
        Run 3 & 0.5200 & 0.4554 & 0.0675 & 0.0346  \\
        \hline
        Run 4 & 0.5050 & 0.4650 & 0.0690 & 0.0353 \\
        \hline
        Run 5 & \textbf{0.5255} & \textbf{0.4690} & \textbf{0.0662} & \textbf{0.0336} \\        
        \bottomrule
    \end{tabular}
\end{table}

\begin{table}[ht]
    \caption{Official results on Japanese data}
    \label{t:dbdc4_result_jp_table}
    \begin{tabular}{ c|c|c|c|c } 
        \toprule
        Run & Accuracy & F1 (B) & JSD (NB, PB, B) & MSE (NB, PB, B) \\
        \midrule
        Run 1 & 0.5390 & 0.4568 & 0.0975 & 0.0492 \\
        \hline
        Run 2 & 0.5412 & \textbf{0.4613} & 0.0989 & 0.0509 \\
        \hline
        Run 3 & \textbf{0.5476} & 0.4589 & 0.0967 & 0.0493  \\
        \hline
        Run 4 & 0.5412 & 0.4583 & 0.0954 & 0.0480 \\
        \hline
        Run 5 & 0.5444 & 0.4603 & \textbf{0.0947} & \textbf{0.0475} \\        
        \bottomrule
    \end{tabular}
\end{table}

\begin{table}[ht]
    \caption{P-value based on Randomised Tukey's HSD test/effect sizes for MSE (NB, PB, B) (English)} 
    \label{t:dbdc4_tukey_mse_en_table}
    \begin{tabular}{ c|c|c|c|c } 
        \toprule
        & Run2 & Run3 & Run4 & Run 5\\
        \midrule
        Run 1 & $p = 0.007 (-0.110)$ & $p < 0.0001 (0.139)$ & $p = 0.0669 (0.080)$ & $p < 0.0001 (0.227)$ \\
        \hline
        Run 2 & - & $p < 0.0001 (0.249)$ & $p < 0.0001 (0.191)$ & $p < 0.0001 (0.337)$ \\
        \hline
        Run 3 & - & - & $p = 0.387 (-0.059)$ & $p = 0.0415 (0.088)$ \\
        \hline
        Run 4 & - & - & - & $p < 0.0001 (0.146)$ \\
        \bottomrule
    \end{tabular}
\end{table}

\begin{table}[ht]
    \caption{P-value based on Randomised Tukey's HSD test/effect sizes for MSE (NB, PB, B) (Japanese)} 
    \label{t:dbdc4_tukey_mse_jp_table}
    \begin{tabular}{ c|c|c|c|c } 
        \toprule
        & Run2 & Run3 & Run4 & Run 5\\
        \midrule
        Run 1 & $p = 0.0086 (-0.104)$ & $p = 1 (-0.002)$ & $p = 0.0338 (0.076)$ & $p < 0.0001 (0.112)$ \\
        \hline
        Run 2 & - & $p = 0.0086 (0.102)$ & $p < 0.0001 (0.181)$ & $p < 0.0001 (0.216)$ \\
        \hline
        Run 3 & - & - & $p = 0.0222 (0.079)$ & $p < 0.0001 (0.114)$ \\
        \hline
        Run 4 & - & - & - & $p = 0.6577 (0.035)$ \\
        \bottomrule
    \end{tabular}
\end{table}

\section{Discussions}
\label{s:discussion}

\subsection{Naive strategy in creating the training data}
\label{s:discuss_training_data}

As described in \cref{s:runs}, in Runs 1-3, we used the same strategy in creating the training data from the given development data. 
For the English submission, we created one group of training data $E_{t1}$ and trained a single model with it. The reason for doing so is that we wanted to create sufficient training data, since there is only a total number of 211 dialogues. 
For the Japanese submission, we created two groups of training data $J_{t1}$ and $J_{t2}$ and trained two models with them respectively. The reason for doing so is that the first group consists of dialogues with 21 turns (fixed to 20 turns in preprocessing) while the second group consists of dialogues with 31 turns (fixed to 30 turns in preprocessing). Because our LSTM-based model only accepts fixed turn lengths, we had to build two models to target two different turn lengths. We used the same strategy for building our Decision Tree-based model so that the ensemble with the LSTM-based model can be done easily.


Nevertheless, the above strategy is rather naive as it does not consider the overall probability distribution of the three breakdown labels for each dialogue system. As shown in \cref{t:dbdc4_en_dev_data_table,t:dbdc4_jp_dev_data_table,t:dbdc4_jp_compe_dev_data_table}, the average probability distribution of each dialogue system is different from one another. In particular, the system IRS in \cref{t:dbdc4_jp_compe_dev_data_table} has a significantly higher probability for label B compared to the other four systems. We believe that IRS should not have been combined with the other four systems to create training data $J_{t2}$. Furthermore, the model which is trained with $J_{t2}$ should not have been used for predicting the data of IRS in $J_{e2}$. 

\cref{t:dbdc4_mse_jp_compe_table} shows the official results of MSE (NB, PB, B) for $J_{e2}$. It can be observed that due to the naive strategy above, all runs achieved poor performance with regard to IRS. To improve the result, we believe that the development data of IRS should be excluded from $J_{t2}$ and combined with $J_{t1}$. When predicting the labels for IRS in $J_{e2}$, we should utilise the model trained with $J_{t1}$ instead of the one trained with $J_{t2}$. This proposed strategy requires us to either fix all training data to 30 turns in the LSTM-based model or develop a new model which accepts a shorter fixed turn length such as 5.

\begin{table}[ht]
    \caption{Official results of MSE (NB, PB, B) for $J_{e2}$}
    \label{t:dbdc4_mse_jp_compe_table}
    \begin{tabular}{ c|c|c|c|c|c } 
        \toprule
         & IRS & MMK & MRK & TRF & ZNK \\
        \midrule
        Run 1 & 0.0662 & 0.0243 & 0.0393 & 0.0282 & 0.0418 \\
        \hline
        Run 2 & 0.0606 & 0.0184 & 0.0328 & 0.0230 & 0.0389 \\
        \hline
        Run 3 & 0.0602 & 0.0195 & 0.0322 & 0.0231 & 0.0394 \\
        \hline
        Run 4 & 0.0606 & 0.0197 & 0.0341 & 0.0236 & 0.0378 \\
        \hline
        Run 5 & 0.0608 & 0.0206 & 0.0340 & 0.0239 & 0.0384 \\
        \bottomrule
    \end{tabular}
\end{table}

\subsection{Ensemble works?}
\label{s:discuss_ensembel}
We analysed our runs in terms of MSE (NB, PB, B) (referred as MSE in this section), which is the emphasised evaluation metric in this challenge. From \cref{t:dbdc4_result_en_table,t:dbdc4_result_jp_table},
it can be observed that Run 4 outperformed Run 1 and Run 2, and Run 5 outperformed Run 1 and Run 3 in terms of MSE for both English and Japanese data.
To investigate how well the ensemble actually worked for each utterance, we would like to know the number of target system utterances for which the ensemble model outperformed the original models that were ensembled.
In this section,
we focus on Run 5 which achieved the best performance in terms of MSE and compare its results with Run 1 and Run 3~\footnote{When comparing the runs in \cref{s:discuss_ensembel}, we remove the first predicted system utterance of every dialogue in Japanese data. This is because the first system utterances in Japanese data are all annotated with the same labels (NB) and are all predicted correctly with MSE = $0.0$ by every run.}.

\cref{t:number_of_turns_en,t:number_of_turns_jp} show the number of target system utterances for which each run outperformed the others. $V$ denotes the set of target system utterances in the evaluation dataset,
and $mse_i(v)$ denotes the MSE of Run $i$ given a target system utterance $v$ ($\in V)$. $V_{1<3,5}$, $V_{3<1,5}$, and $V_{5<1,3}$ are defined by the following equations:

\begin{equation}
    V_{1<3,5} = \{v | mse_1(v) < mse_5(v) < mse_3(v), v \in V \},
\end{equation}
\begin{equation}
    V_{3<1,5} = \{v | mse_3(v) < mse_5(v) < mse_1(v), v \in V \},
\end{equation}
\begin{equation}
    V_{5<1,3} = \{v | mse_5(v) < mse_1(v) \land　mse_5(v) < mse_3(v), v \in V \},
\end{equation}

\setlength{\tabcolsep}{3mm}

\begin{table}[h!]
    \caption{Number of turns for which each Run outperformed the others for the English data}
    \label{t:number_of_turns_en}
    \begin{tabular}{l|r}
        \toprule
        a subset of turns  $V'$ ($\subset V$) & $|V'|$ \\
        \midrule
        $V_{1<3,5}$ & 866 \\
        $V_{3<1,5}$ & 958 \\
        $V_{5<1,3}$ & 176 \\
        \midrule
        $\{v | mse_1(v) < mse_5(v) \land　mse_3(v) < mse_5(v), v \in V \}$ & 0 \\
        \bottomrule
    \end{tabular}
\end{table}

\begin{table}[h!]
    \caption{Number of turns for which each Run outperformed the others for the Japanese data}
    \label{t:number_of_turns_jp}
    \begin{tabular}{l|r}
        \toprule
        a subset of turns  $V'$ ($\subset V$) & $|V'|$ \\
        \midrule
        $V_{1<3,5}$ & 1200 \\
        $V_{3<1,5}$ & 1233 \\
        $V_{5<1,3}$ & 162 \\
        \midrule
        $\{v | mse_1(v) < mse_5(v) \land　mse_3(v) < mse_5(v), v \in V \}$ & 0 \\
        \bottomrule
    \end{tabular}
\end{table}

\begin{figure}[h!]
  \includegraphics[width=110mm]{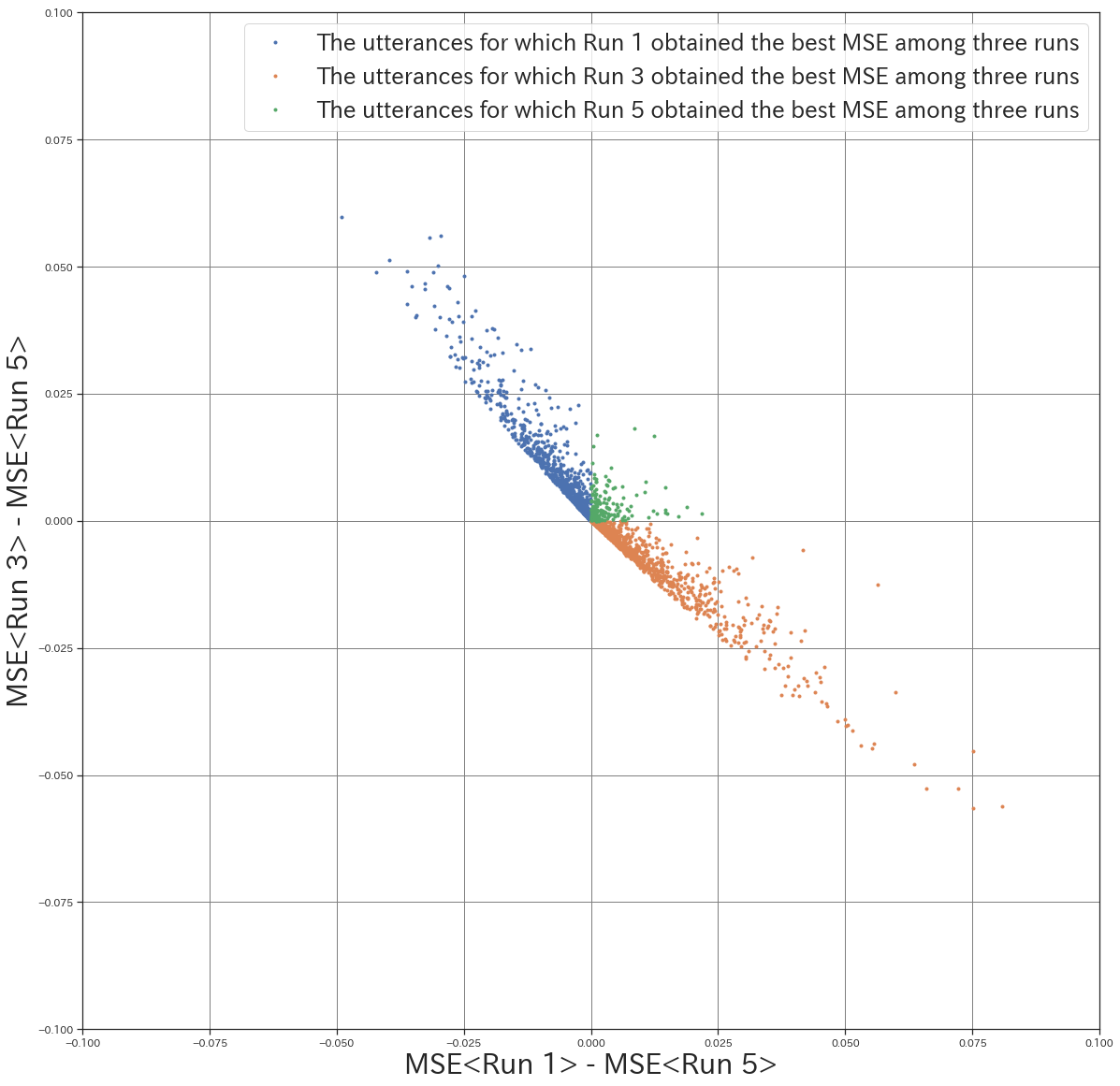}
  \caption{Relationship of the differences between the MSE of Run 1, Run 3 and Run 5 for the English data}
  \label{f:diff_and_diff_en}
\end{figure}

\begin{figure}[h!]
  \includegraphics[width=110mm]{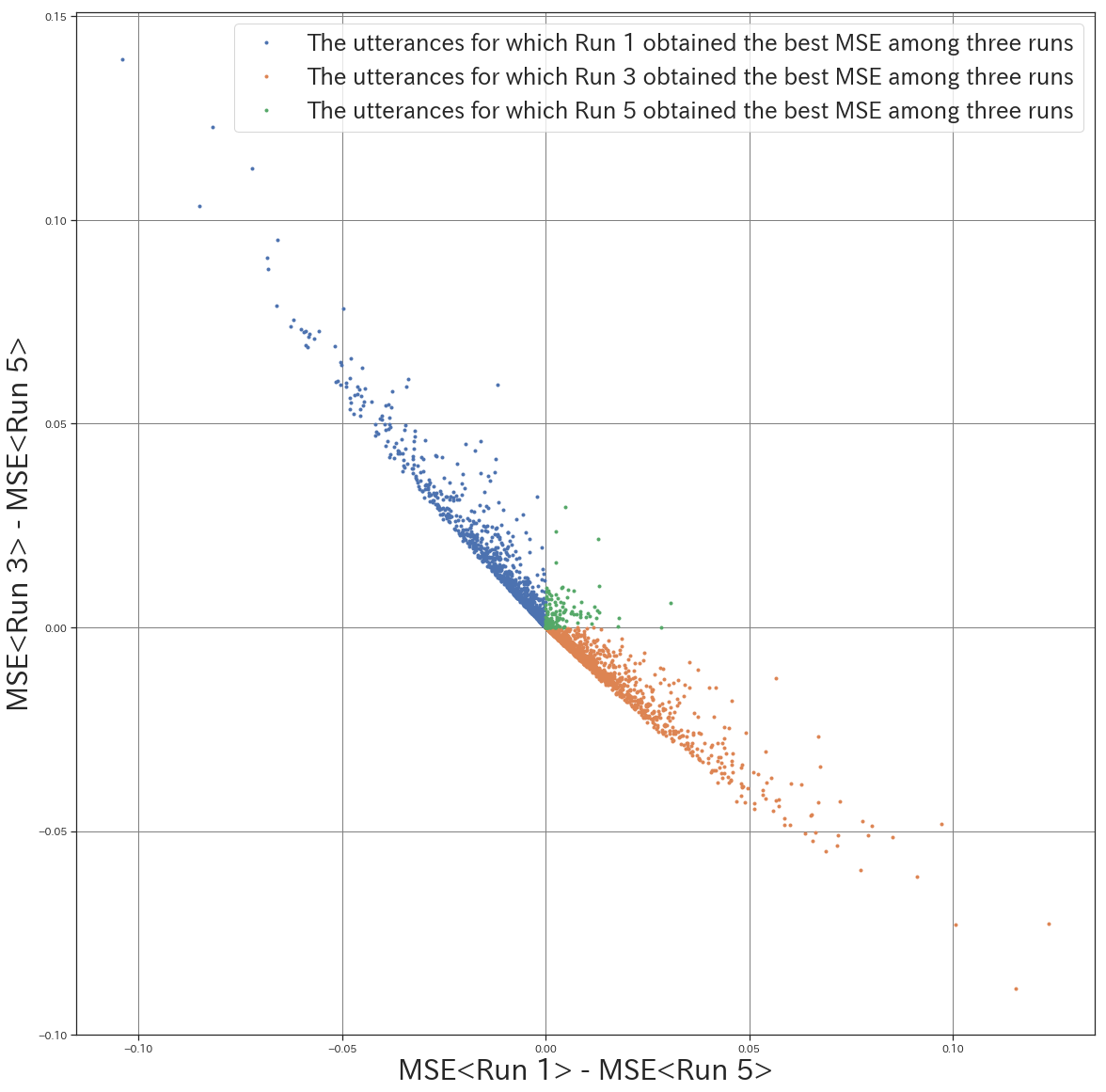}
  \caption{Relationship of the differences between the MSE of Run 1, Run 3 and Run 5 for the Japanese data}
  \label{f:diff_and_diff_jp}
\end{figure}

From \cref{t:number_of_turns_en,t:number_of_turns_jp},
it can be observed that
the number of target system utterances for which Run 5 outperformed the other runs is relatively small.
We plotted the relationship of the differences between the MSE of Run 1, Run 3, and Run 5 in \cref{f:diff_and_diff_en,f:diff_and_diff_jp}. The x-axis is $mse_1(v) - mse_5(v)$, and the y-axis is $mse_3(v) - mse_5(v)$. The points coloured in blue, orange, and green denote the target system utterances that match the condition of $V_{1<3,5}$, $V_{3<1,5}$, and $V_{5<1,3}$ respectively.

By observing \cref{f:diff_and_diff_en,f:diff_and_diff_jp},
it appears that the condition which makes the MSE of Run 5 lower than the ones of Run 1 and Run 3 is that the target system utterance is located in the first quadrant of \cref{f:diff_and_diff_en,f:diff_and_diff_jp}.

\begin{figure}[h!]
  \includegraphics[width=110mm]{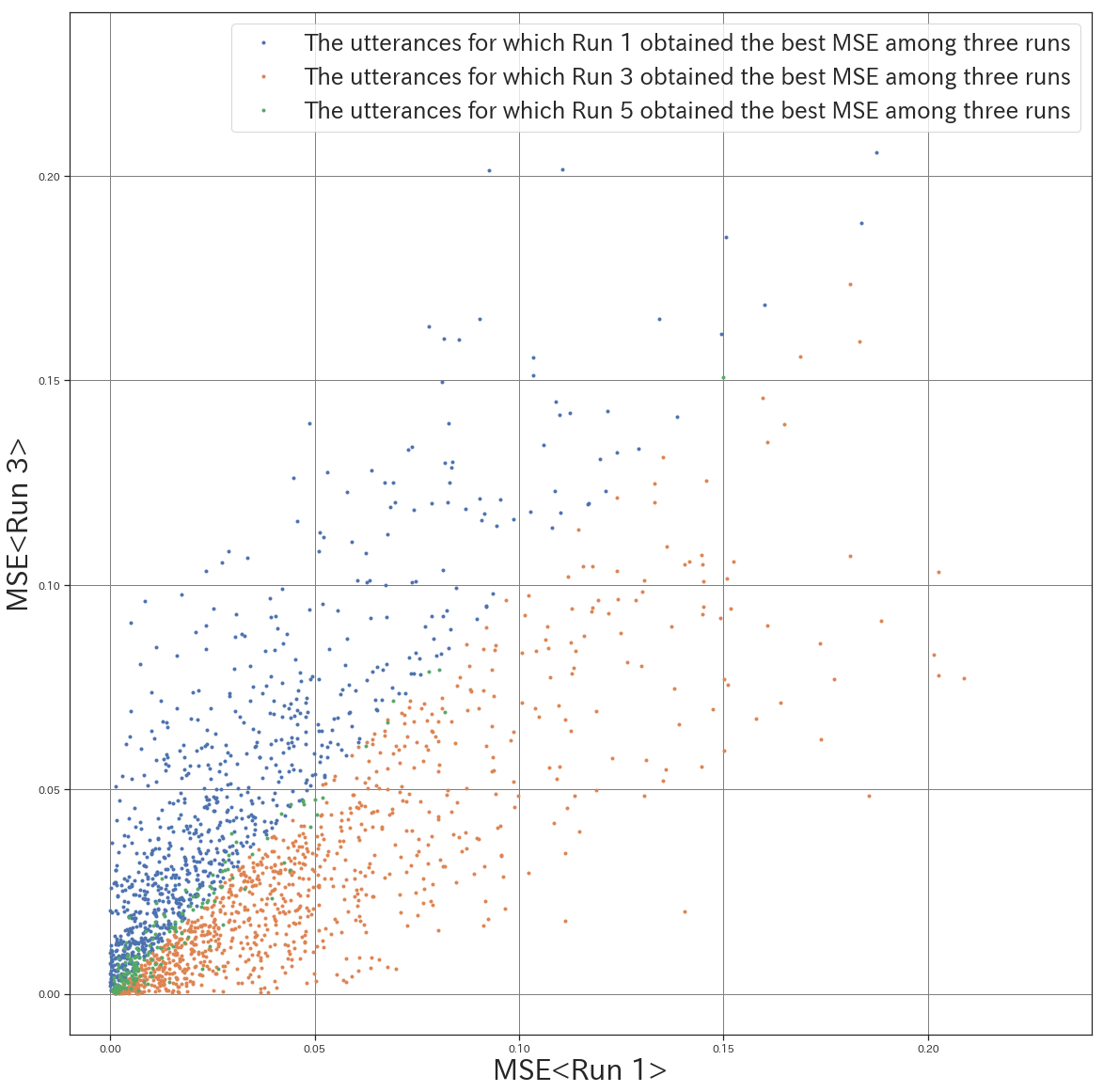}
  \caption{Relationship of MSEs between Run 1 and Run 3 for the English data}
  \label{f:1_and_3_en}
\end{figure}

\begin{figure}[h!]
  \includegraphics[width=110mm]{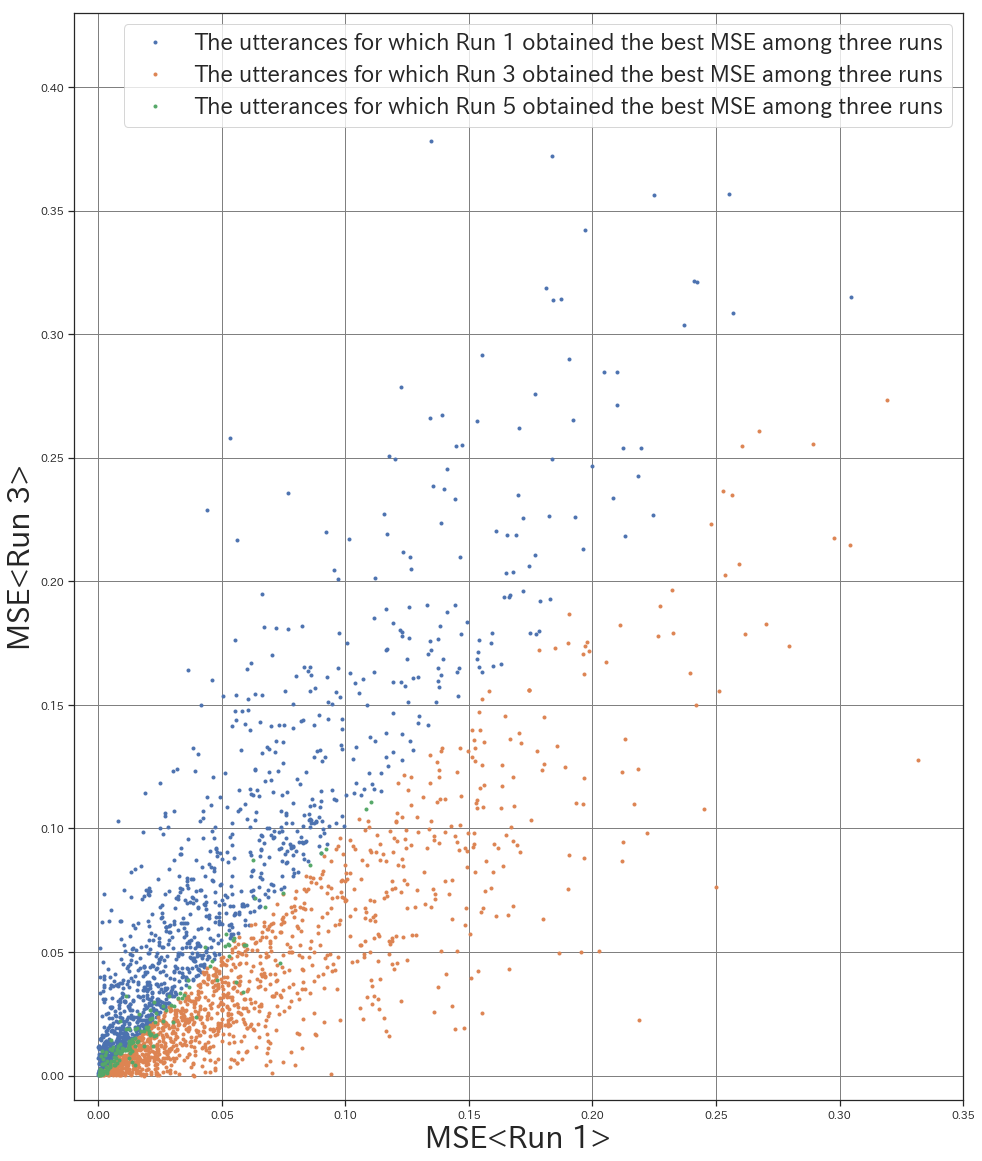}
  \caption{Relationship of MSEs between Run 1 and Run 3 for the Japanese data}
  \label{f:1_and_3_jp}
\end{figure}

\cref{t:mean_MSE_over_subset_en,t:mean_MSE_over_subset_jp} show the mean MSE of Run 1, Run 3, and Run 5 over $V_{1<3,5}$, $V_{3<1,5}$, and $V_{5<1,3}$ respectively. From \cref{t:mean_MSE_over_subset_en,t:mean_MSE_over_subset_jp},
it can be observed that when Run 5 outperformed Run 1 and Run 3, the MSEs of Run 1 and Run 3 tend to be low. Similar to \cref{f:diff_and_diff_en,f:diff_and_diff_jp},
we plotted the relationship of the MSE between Run 1 and Run 3 in \cref{f:1_and_3_en,f:1_and_3_jp}.

\begin{table}[h!]
    \caption{Mean MSE over $V_{1<3,5}$, $V_{3<1,5}$ and $V_{5<1,3}$ for the English data}
    \label{t:mean_MSE_over_subset_en}
    \begin{tabular}{c|rrr}
        \toprule
        a subset of turns  $V'$ ($\subset V$) & Run 1 & Run 3 & Run 5 \\
        \midrule
        $V_{1<3,5}$ & 0.0270 & 0.0451 & 0.0344 \\
        $V_{3<1,5}$ & 0.0481 & 0.0285 & 0.0367 \\
        $V_{5<1,3}$ & 0.0159 & 0.0159 & 0.0129 \\
        \bottomrule
    \end{tabular}
\end{table}

\begin{table}[h!]
    \caption{Mean MSE over $V_{1<3,5}$, $V_{3<1,5}$ and $V_{5<1,3}$ for the Japanese data}
    \label{t:mean_MSE_over_subset_jp}
    \begin{tabular}{c|rrr}
        \toprule
        a subset of turns  $V'$ ($\subset V$) & Run 1 & Run 3 & Run 5 \\
        \midrule
        $V_{1<3,5}$ & 0.0463 & 0.0721 & 0.0573 \\
        $V_{3<1,5}$ & 0.0649 & 0.0399 & 0.0505 \\
        $V_{5<1,3}$ & 0.0195 & 0.0194 & 0.0164 \\
        \bottomrule
    \end{tabular}
\end{table}

\begin{figure}[h!]
  \includegraphics[width=110mm]{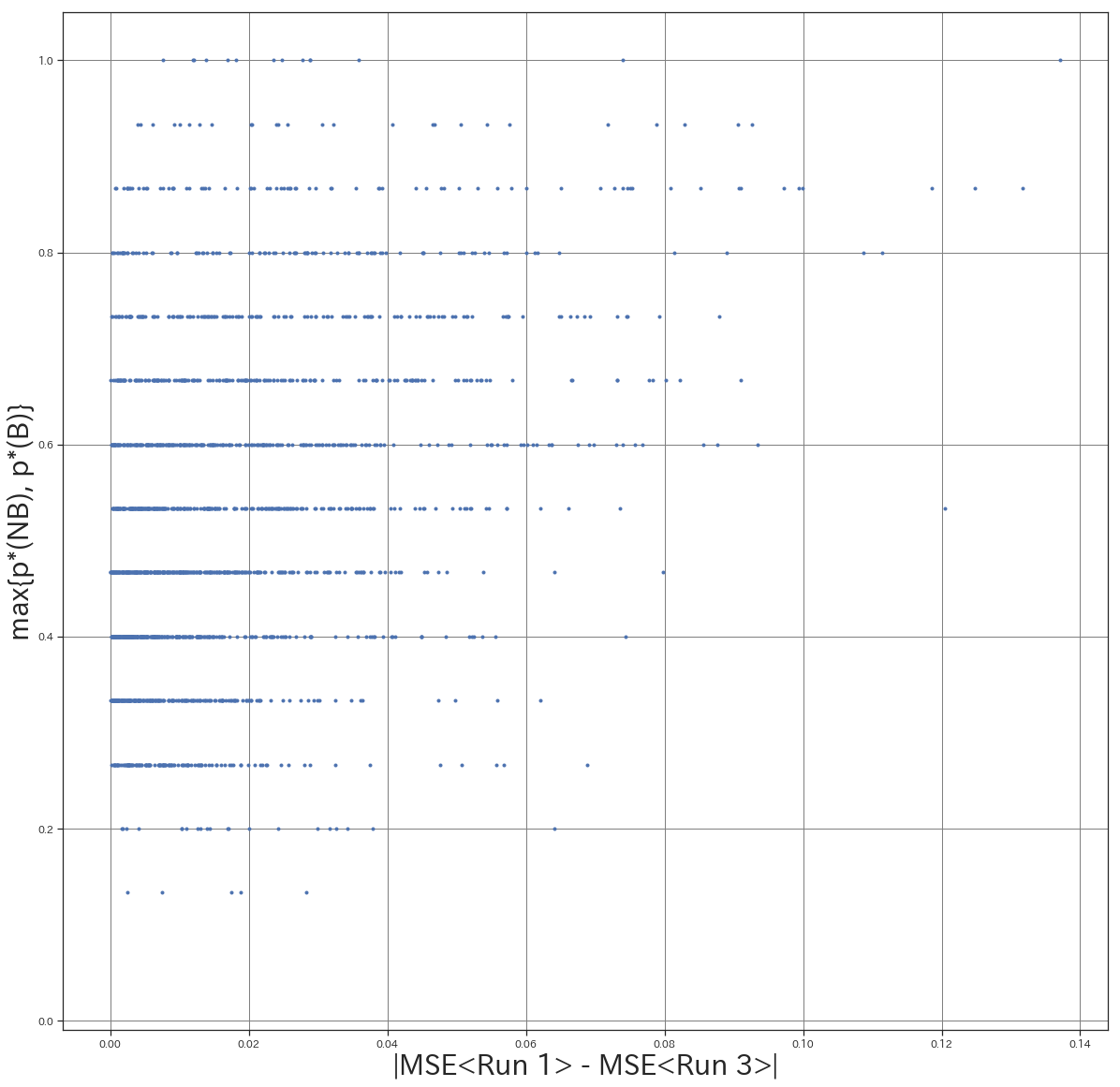}
  \caption{Relationship of the absolute difference between the MSE of Run 1 and Run 3 and $\max \{p^*(NB), p^*(B)\}$ for the English data}
  \label{f:diff_and_max_en}
\end{figure}

\begin{figure}[h!]
  \includegraphics[width=110mm]{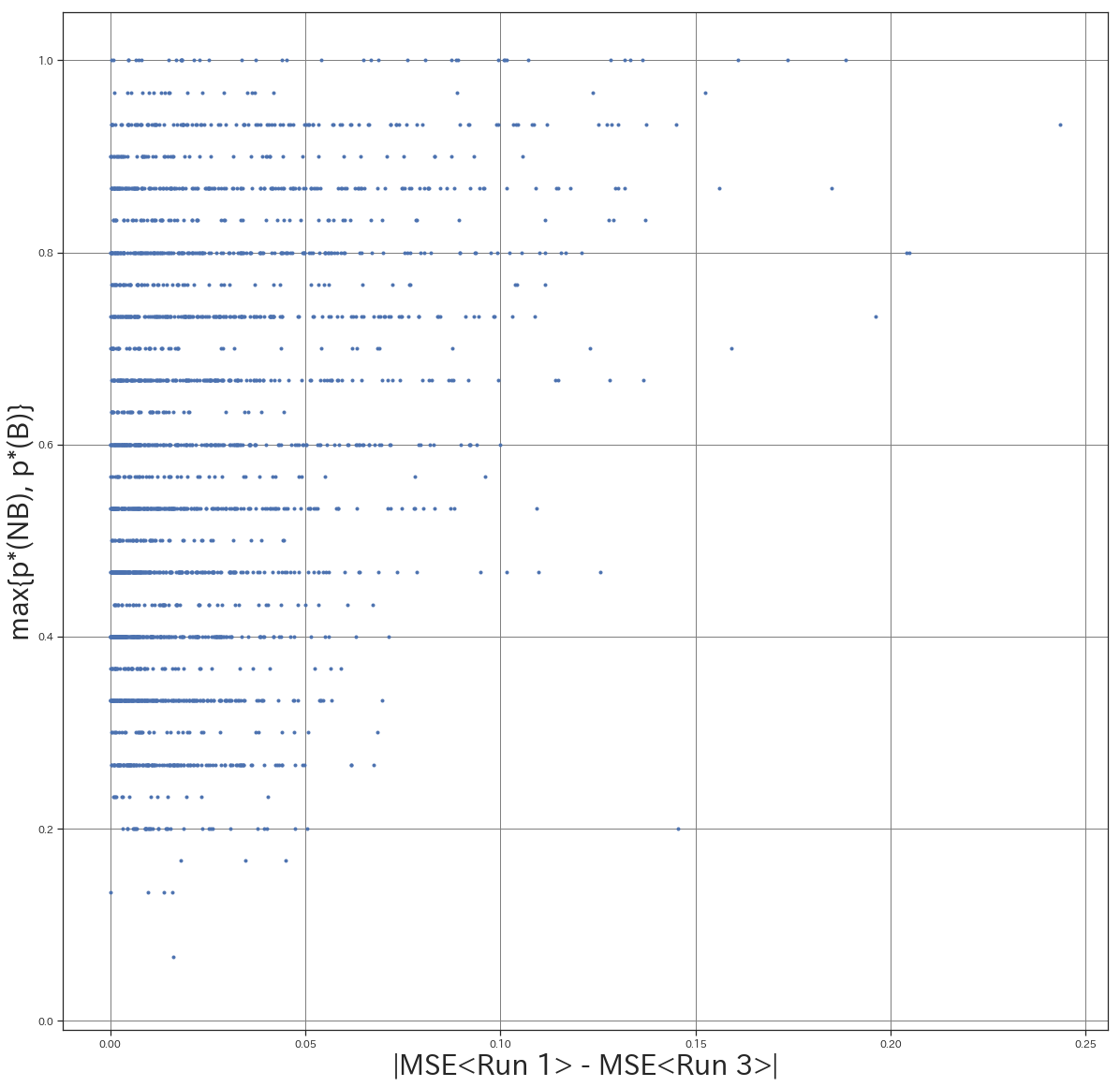}
  \caption{Relationship of the absolute difference between the MSE of Run 1 and Run 3 and $\max \{p^*(NB), p^*(B)\}$ for the Japanese data}
  \label{f:diff_and_max_jp}
\end{figure}

By observing \cref{f:1_and_3_en,f:1_and_3_jp}, it appears that the green points are concentrated at the origin of both axes. In addition, Run 5 tends to outperform the two other runs when the MSE of Run 1 and Run 3 are similar. We looked into the system utterances for which the difference between the MSE of Run 1 and Run 3 are high and found out these utterances tend to be labeled with high probability of NB or B compared to other utterances. We plotted the relationship of the absolute difference between the MSE of Run 1 and Run 3 and $\max \{p^*(NB), p^*(B)\}$ in \cref{f:diff_and_max_en,f:diff_and_max_jp}, where $\max \{p^*(NB), p^*(B)\}$ denotes the maximum probability of the labeled probabilities of NB and B. The points coloured in blue are the target system utterances.

From \cref{f:diff_and_max_en,f:diff_and_max_jp}, it can be observed that the MSE of Run 1 and Run 3 tend to be similar when $\max \{p^*(NB), p^*(B)\}$ is low. This means that the ensemble model tends to perform the best in target system utterances which are not labeled with high probability of NB or B. Therefore, to further improve our ensemble model, we should either develop a new ensemble strategy different from simple averaging or include a third model which focuses on minimising the MSE in target system utterances that are labeled with high probability of NB or B.

\clearpage

\section{Conclusions}
\label{s:conclusion}
We submitted five runs to both English and
Japanese subtasks of DBDC4. Run 1 utilises a Decision Tree-based model; Run 2 utilises an LSTM-based model; Run 3 performs an ensemble of 5 LSTM-based models; Run 4 performs an ensemble of Run 1 and Run 2; Run 5 performs an ensemble of Run 1 and Run 3. Run 5 statistically significantly outperformed all other runs in terms of MSE (NB, PB, B) for the English data and all other runs except Run 4 in terms of MSE (NB, PB, B) for the Japanese data (alpha level = 0.05).

Our future work includes utilising a proposed strategy in creating the training data and improving our ensemble model. The proposed strategy considers the overall probability distribution of the three breakdown labels for each dialogue systems and requires us to either fix all training data to 30 turns in the LSTM-based model or develop a new model which accepts a shorter fixed turn length such as 5. To improve our ensemble model, we should either develop a new ensemble strategy different from simple averaging or include a third model which focuses on minimising the MSE in target system utterances that are labeled with high probability of NB or B.




\bibliographystyle{spmpsci}
\bibliography{references}

\end{document}